\documentclass[10pt,letterpaper,twocolumn]{article}
\usepackage[T1]{fontenc}
\usepackage{mathptmx}
\usepackage[margin=0.75in,columnsep=0.375in]{geometry}
\usepackage{placeins}
\usepackage{authblk}
\usepackage{hyperref}
\hypersetup{hidelinks,pdftitle={ViTok},pdfauthor={Hailun Xu and Kanchan Sarkar}}
\usepackage{url}
\usepackage{graphicx}
\usepackage{natbib}
\usepackage{caption}
\usepackage{amsmath}
\usepackage{amssymb}
\usepackage{bm}
\usepackage{booktabs}
\usepackage{array}
\usepackage{multirow}
\usepackage{nicefrac}

\newcommand{\method}{\textsc{ViTok}}
\newcommand{\baseline}{AM-RADIO}
\newcommand{\phis}{PHI-S}
\newcommand{\siglip}{SigLIP2}
\newcommand{\dinothree}{DINOv3-L}

\title{\textbf{ViTok: Improving Dense Semantics in AM-RADIO-Style Multi-Teacher Distillation with PHI-S and Masked Image Modeling}}
\author[1]{Hailun Xu}
\author[2]{Kanchan Sarkar\thanks{Corresponding author: \href{mailto:kanchansarkar9975@iitbombay.org}{kanchansarkar9975@iitbombay.org}.}}
\affil[1]{Beihang University}
\affil[2]{Indian Institute of Technology}
\date{}

\begin{document}

\maketitle

\begin{abstract}
We study how to consolidate the current \method{} progress into a single multi-teacher distillation recipe that jointly preserves global recognition and dense semantics. Our starting point is an \baseline{}-style student distilled from \siglip{} and \dinothree{}, where \siglip{} supplies strong global semantics and \dinothree{} supplies stronger dense features. The central empirical issue is that the same recipe does not optimize all objectives equally well: changes that improve ImageNet-1K kNN accuracy can still degrade ADE20K segmentation. We summarize a progression of modifications that make this trade-off more explicit and more manageable: split adaptor heads for CLS and patch tokens, asymmetric cosine/MSE losses, initialization from a \dinothree{} checkpoint, teacher reweighting, masked image modeling (MIM), and \phis{} feature balancing. The resulting model reaches 83.2 patch kNN and 85.2 CLS kNN, slightly surpassing the \dinothree{} teacher on ImageNet-1K kNN classification, while \phis{} restores ADE20K performance from 46.5/58.1 to 48.5/61.0 mIoU/mAcc, matching the teacher on this dense benchmark. We also summarize negative results: scaling distillation from ImageNet-1K to ImageNet-22K does not consistently help, and naively adding extra teachers such as SAM3 or HOG features introduces interference. Rather than claiming a final recipe, this paper distills the current project state into a compact empirical story and a concrete set of lessons for future iterations.

\end{abstract}

\section{Introduction}

Modern vision encoders are highly specialized. Vision-language models such as \siglip{} provide strong global semantics and image classification performance, while self-distilled models such as \dinothree{} typically produce stronger dense features for segmentation and other spatial tasks. This specialization motivates multi-teacher distillation: instead of selecting a single teacher, we can aggregate complementary capabilities into one student. \baseline{}~\citep{ranzinger2024amradio} showed that this agglomerative view is practical, but our current project reveals a harder question: how can a single student retain both strong global recognition and strong dense semantics when the teachers have mismatched feature statistics?

\begin{figure*}[t]
  \centering
  \includegraphics[width=0.8\textwidth]{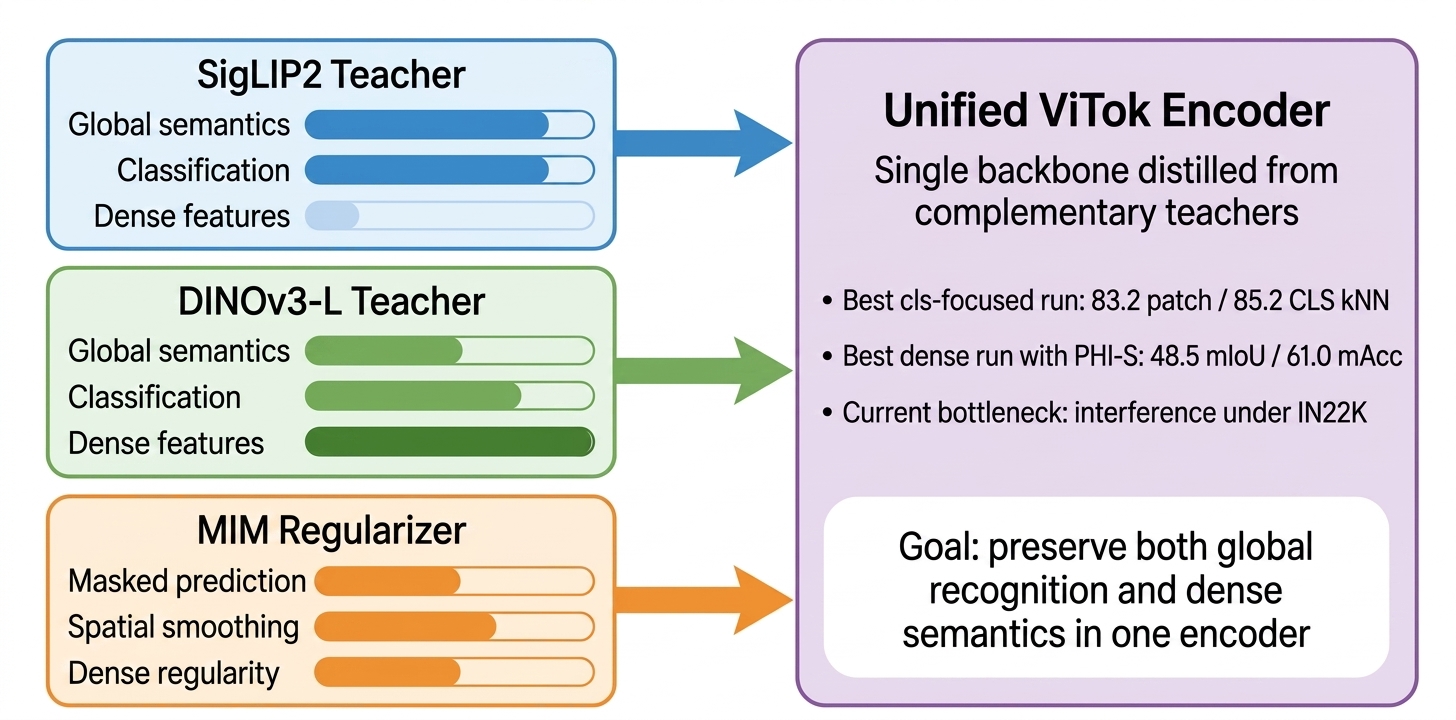}
  \caption{Motivation for \method{}. We treat \siglip{} and \dinothree{} as complementary teachers and use MIM as a dense regularizer. The goal is to recover a more balanced encoder that maintains both classification and dense prediction quality.}
  \label{fig:motivation}
\end{figure*}

Our project, which we refer to as \method{}, focuses on this tension in the concrete setting of \siglip{}+\dinothree{} distillation. The progress log shows a clear pattern. Structural changes such as split adaptors and asymmetric losses improve ImageNet-1K kNN classification; initialization from \dinothree{} and teacher reweighting further improve the classification side; however, ADE20K segmentation still lags unless feature balancing is introduced. This is where \phis{}~\citep{ranzinger2024phis} becomes critical: by normalizing teacher features in a rotation-aware way, it recovers dense performance that would otherwise be lost.

This paper is a progress-oriented consolidation rather than a final, fully closed story. We intentionally summarize stable observations and separate them from open issues. Our current conclusions are threefold. First, the adaptor and loss design matters: split CLS/patch heads with cosine loss for CLS tokens and MSE for patch tokens provide a much better classification operating point than a single adaptor. Second, dense semantics are fragile under multi-teacher distillation, and \phis{} is the main ingredient that restores them. Third, larger-scale distillation and additional teachers are not automatically helpful; ImageNet-22K, SAM3, and HOG-based extensions all expose interference that remains unresolved.

The remainder of the paper follows this progression. We first place \method{} in the context of agglomerative distillation and masked pretraining. We then summarize the \baseline{} formulation, introduce the \method{} modifications and a compact derivation of the resulting training objective, and finally present the current empirical picture: classification improves through structure and weighting, dense prediction recovers through \phis{}, and scaling this two-teacher recipe further remains difficult.

\section{Related Work}

\paragraph{Vision Transformers and dense representations.} ViT~\citep{dosovitskiy2021vit} established patch-based transformers as a scalable alternative to convolutional backbones by treating an image as a sequence of non-overlapping patches. Later variants such as DeiT~\citep{touvron2021deit} and Swin Transformer~\citep{liu2021swin} improved data efficiency and introduced hierarchical structure useful for detection and segmentation. While hierarchical backbones like Swin make dense transfer easier by construction, recent foundation encoders increasingly rely on plain ViT-style architectures whose dense behavior emerges from scale, self-supervision, or auxiliary tasks. Our student falls in this latter category: it uses a single transformer encoder but must preserve both summary-level semantics and patch-level structure.

\paragraph{Self-supervised transformer features.} Self-distillation has made ViT features much stronger for dense perception. DINO~\citep{caron2021dino} first revealed strong emergent properties in self-supervised vision transformers, and \citet{chen2021selfsupervisedvit} showed through MoCo v3 that transformer self-supervision is unusually sensitive to optimization details. iBOT~\citep{zhou2022ibot} then coupled CLS-level semantics and patch-token learning via masked prediction with an online tokenizer. DINOv2~\citep{oquab2024dinov2} and \dinothree{}~\citep{dinov32024report} pushed this line into large-scale visual representations with strong localization and segmentation behavior. These models are directly relevant because our project inherits both their strength and their tension: DINO-style features are excellent dense teachers, but they are also the ones most easily degraded when mixed with strong global-semantic teachers. Understanding the optimization sensitivity documented in these works informs our choice of learning-rate schedules and the care we take when introducing auxiliary objectives such as MIM.

\paragraph{Vision-language supervision and multi-teacher distillation.} CLIP~\citep{radford2021clip}, SigLIP~\citep{zhai2023siglip}, and \siglip{}~\citep{zhai2024siglip2} provide strong global semantics and open-vocabulary alignment through image-text supervision. Systems like SLIP~\citep{mu2022slip} and GLIP~\citep{li2022glip} show that language-image and grounding-oriented supervision can complement self-supervised features. In our setting these models matter less as direct baselines than as sources of supervision whose inductive biases differ from dense self-supervised teachers. Recent general-purpose vision encoders increasingly combine such heterogeneous signals: \baseline{}~\citep{ranzinger2024amradio} made this aggregation viewpoint explicit by jointly mimicking CLIP-like global semantics, DINO-style dense features, and SAM~\citep{kirillov2023sam} segmentation priors through teacher-specific heads. Our work follows this line but focuses on the narrower \siglip{}+\dinothree{} setting where the main difficulty is not teacher coverage but feature interference between global and dense objectives. By restricting the teacher set we can study this interference in isolation and identify interventions---split heads, \phis{} normalization, and auxiliary MIM---that specifically target the global-dense conflict.

\paragraph{Feature normalization and masked modeling.} Feature normalization has emerged as a central ingredient in multi-teacher distillation. \phis{}~\citep{ranzinger2024phis} shows that naive MSE or Smooth-L1 matching implicitly overweights high-variance teachers, and proposes a Hadamard-rotated normalization that equalizes variance more isotropically. We adopt its intuition to regularize teacher targets and also test a variant that computes separate statistics for CLS and patch tokens. On the masked modeling side, MAE~\citep{he2022mae} and BEIT~\citep{bao2021beit} establish standard reconstruction objectives, EVA~\citep{fang2023eva} replaces pixel targets with CLIP-aligned features, and MaskFeat~\citep{wei2022maskfeat} shows that masked prediction over hand-crafted descriptors also produces strong features. Taken together with iBOT, these methods suggest that patch-level supervision is most useful when it preserves semantic structure. Our project adopts this intuition in a constrained way: MIM is not stand-alone pretraining but an auxiliary regularizer inside multi-teacher distillation to stabilize dense structure without discarding the benefits of \siglip{}. Importantly, our ablations reveal that this auxiliary role is conditional---MIM can destabilize training under certain learning-rate regimes---underscoring the need for careful co-design of auxiliary objectives and optimization settings in multi-teacher pipelines.

\section{Preliminaries}
\label{sec:prelim}

We adopt the \baseline{} view of agglomerative distillation. Let a student encoder $f_\theta$ produce a sequence of latent tokens
\begin{equation}
  x = f_\theta(I) = [x^{\mathrm{cls}}, x^{\mathrm{patch}}],
\end{equation}
where $I$ is an input image, $x^{\mathrm{cls}} \in \mathbb{R}^{C}$ is the summary token, and $x^{\mathrm{patch}} \in \mathbb{R}^{N \times C}$ are patch tokens. For each teacher $t \in \mathcal{T}$, we use teacher-specific adaptor heads
\begin{equation}
  h_t^{\mathrm{cls}}: \mathbb{R}^{C} \rightarrow \mathbb{R}^{C_t},
  \qquad
  h_t^{\mathrm{patch}}: \mathbb{R}^{N \times C} \rightarrow \mathbb{R}^{N_t \times C_t},
\end{equation}
to map student features into the teacher space.

Each teacher produces a summary target $z_t^{\mathrm{cls}}$ and a patch target $z_t^{\mathrm{patch}}$. In the original \baseline{} framing, the student matches both kinds of targets with teacher-specific projection heads and aggregates the resulting losses across teachers~\citep{ranzinger2024amradio}. In our project, the active teacher set is usually $\mathcal{T}=\{\text{\siglip{}}, \text{\dinothree{}}\}$, while later explorations additionally test SAM3 and HOG-style features.

The project log also motivates a distinction between global and dense behavior. The summary token is most tightly coupled to image-level recognition, so its quality is measured with CLS-token kNN on ImageNet-1K. Patch tokens are used both for patch-token kNN and for downstream segmentation, so they are the main carrier of dense semantics. This separation is the reason our modifications consistently distinguish CLS and patch tokens rather than treating all tokens identically.

\section{Method}
\label{sec:method}

\subsection{Split adaptors and asymmetric token losses}

Our first modification to the \baseline{} recipe is structural. Instead of using a single adaptor for all tokens, we use separate adaptor heads for CLS and patch tokens. This allows the student to align global and dense teacher targets with different inductive biases. The summary branch uses cosine distance,
\begin{equation}
  \mathcal{L}_{\mathrm{cls}}
  =
  \sum_{t \in \mathcal{T}} \lambda_t^{\mathrm{cls}}
  \left(
    1 - \cos\left(h_t^{\mathrm{cls}}(x^{\mathrm{cls}}), \tilde{z}_t^{\mathrm{cls}}\right)
  \right),
  \label{eq:lcls}
\end{equation}
while the patch branch uses MSE,
\begin{equation}
  \mathcal{L}_{\mathrm{patch}}
  =
  \sum_{t \in \mathcal{T}} \lambda_t^{\mathrm{patch}}
  \left\|
    h_t^{\mathrm{patch}}(x^{\mathrm{patch}}) - \tilde{z}_t^{\mathrm{patch}}
  \right\|_2^2.
  \label{eq:lpatch}
\end{equation}
The asymmetry is empirical rather than theoretical: cosine loss worked better for the CLS branch, whereas MSE preserved patch-token quality more effectively in our progress log.

\begin{figure*}[t]
  \centering
  \includegraphics[width=0.8\textwidth]{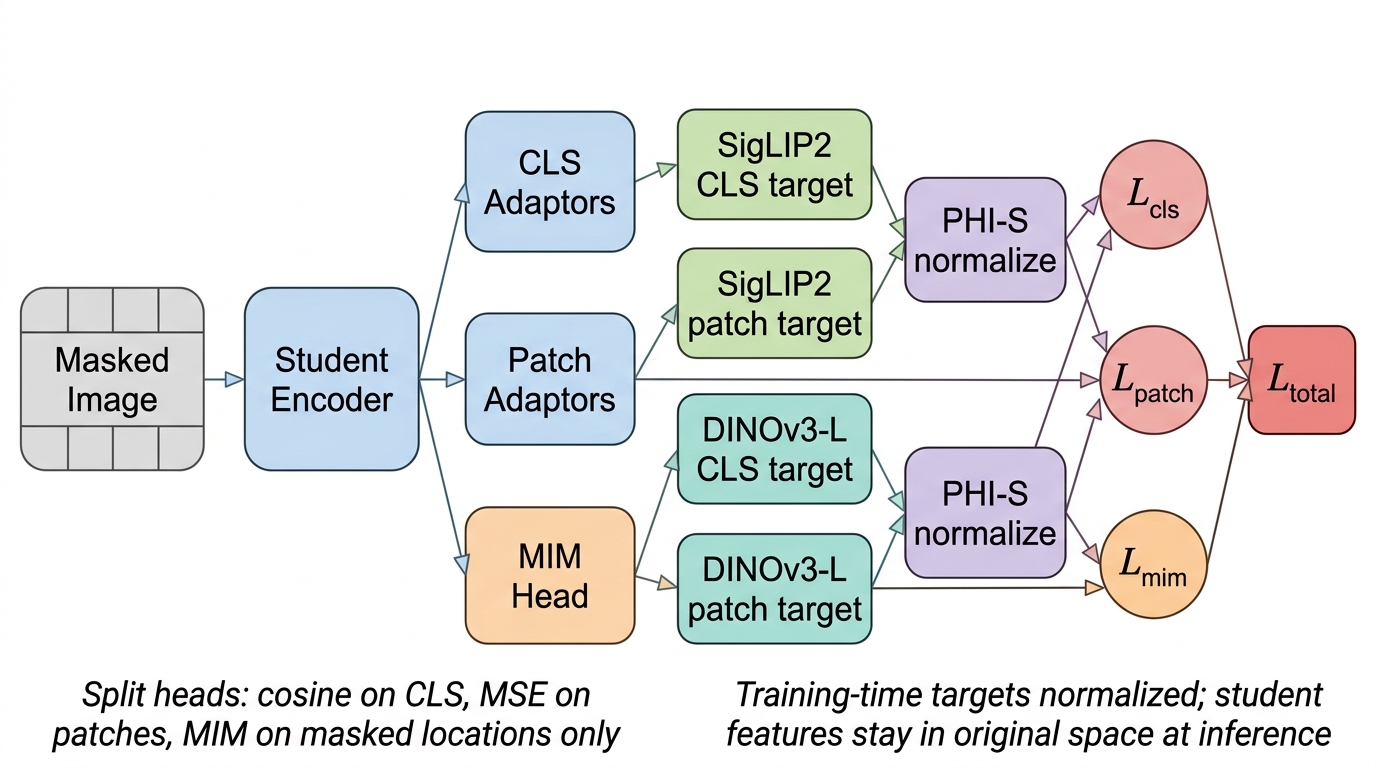}
  \caption{Overview of the current \method{} recipe. The student encoder receives a masked image, sends CLS and patch tokens through separate teacher-specific adaptor heads, matches normalized teacher targets, and optimizes an auxiliary MIM loss on masked locations. PHI-S is applied only during target construction; inference remains in the original student feature space.}
  \label{fig:method}
\end{figure*}

\subsection{PHI-S target balancing}

Feature balancing follows the central insight of \phis{}~\citep{ranzinger2024phis}: when different teachers have very different activation scales, unnormalized MSE implicitly changes their effective weights. Let $z_t$ denote a teacher feature, with mean $\mu_t$ and covariance
\begin{equation}
  \Sigma_t = U_t \Lambda_t U_t^\top,
\end{equation}
where $\Lambda_t = \mathrm{diag}(\lambda_{t,1}, \dots, \lambda_{t,C_t})$. The \phis{} transform rotates the teacher distribution and uses a single isotropic scale:
\begin{equation}
  \tilde{z}_t = W_t (z_t - \mu_t),
  \qquad
  W_t = \alpha_t H_t U_t^\top,
  \label{eq:phistarget}
\end{equation}
where $H_t$ is a normalized Hadamard matrix and
\begin{equation}
  \alpha_t
  =
  \left(
    \frac{1}{C_t}
    \sum_{i=1}^{C_t} \lambda_{t,i}
  \right)^{-\nicefrac{1}{2}}.
  \label{eq:alpha}
\end{equation}
Equation~\eqref{eq:alpha} follows from the requirement that, after rotation by $H_t U_t^\top$, each channel share the same average variance, so a single scalar normalization is sufficient to make the rotated channels unit-scale on average. We use this transform only for building training targets. The student itself remains in its native feature space.

In the later ImageNet-22K stage, we also test a project-specific variant that computes separate \phis{} statistics for CLS and patch tokens. This extension is easy to write down with separate pairs $(W_t^{\mathrm{cls}}, W_t^{\mathrm{patch}})$, but empirically it did not outperform the simpler shared-token formulation.

\subsection{Masked image modeling and the full objective}

We further add masked image modeling as an auxiliary dense regularizer. Let $\mathcal{M}$ denote the masked patch set, and let $g_\phi$ be the prediction head used for masked locations. We write the MIM term in a generic reconstruction-style form,
\begin{equation}
  \mathcal{L}_{\mathrm{mim}}
  =
  \frac{1}{|\mathcal{M}|}
  \sum_{i \in \mathcal{M}}
  \left\|
    g_\phi(x_i^{\mathrm{patch}}) - q_i
  \right\|_2^2,
  \label{eq:lmim}
\end{equation}
where $q_i$ denotes the masked-token target used in the corresponding experiment. The progress log records mask ratios and MIM weights more reliably than the final target design, so we keep this term intentionally generic and treat it as a dense auxiliary objective.

The resulting training objective is
\begin{equation}
  \mathcal{L}_{\mathrm{total}}
  =
  \mathcal{L}_{\mathrm{cls}}
  +
  \mathcal{L}_{\mathrm{patch}}
  +
  \lambda_{\mathrm{mim}} \mathcal{L}_{\mathrm{mim}}.
  \label{eq:ltotal}
\end{equation}
Relative to prior \baseline{}/\phis{}-style distillation, the concrete recipe studied here is therefore defined by six changes: split CLS/patch adaptors, asymmetric losses, \dinothree{} initialization, explicit teacher reweighting, auxiliary MIM, and \phis{} feature balancing.

\begin{figure}[t]
  \centering
  \includegraphics[width=0.95\columnwidth]{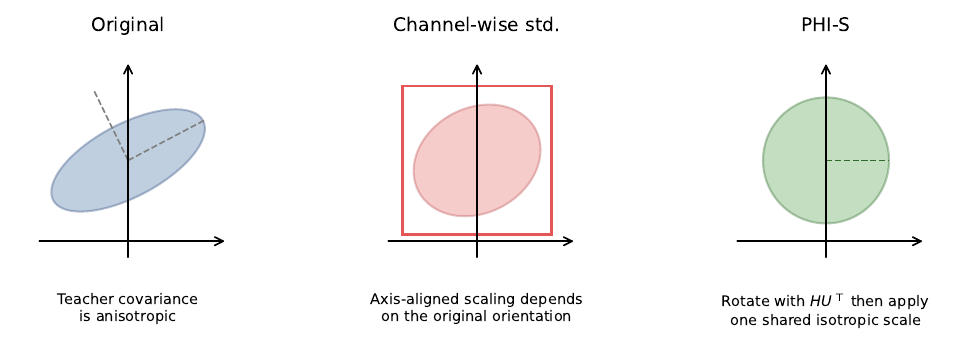}
  \caption{Geometry of PHI-S. Regular channel-wise standardization depends on axis alignment and can overemphasize low-variance directions. PHI-S first rotates the covariance basis and then applies a single isotropic scale, which better preserves balanced dense behavior in our experiments.}
  \label{fig:phis}
\end{figure}

\section{Experiments}
\label{sec:experiments}

\paragraph{Datasets and evaluation.}
The progress log covers distillation on ImageNet-1K and ImageNet-22K, kNN evaluation on ImageNet-1K validation, and linear semantic segmentation on ADE20K. We report both patch-token and CLS-token kNN accuracy because the project consistently observes different behavior on the two token types. For dense prediction, we report ADE20K mIoU and mAcc.

\paragraph{Teacher configuration.}
The main recipe uses \siglip{} and \dinothree{} as teachers. This pairing is the most stable combination in the project log and is the core system throughout the paper. Later experiments add SAM3 and HOG-like targets to probe whether stronger dense supervision can help; these runs are treated as exploratory rather than final.

\paragraph{Training variants.}
Across the ablations, we vary the adaptor structure, token-wise loss assignment, initialization, teacher weights, MIM, \phis{}, training dataset, and the optional extra teachers. We use a compact shorthand for teacher weighting: \texttt{sXdY} means \texttt{SigLIP loss weight = X} and \texttt{DINO loss weight = Y}. For example, \texttt{s1d1} means both teacher losses are weighted by $1$, \texttt{s1d0.5} means SigLIP has weight $1$ and DINO has weight $0.5$, and \texttt{s0.5d1} means SigLIP has weight $0.5$ and DINO has weight $1$. Not every run logs every hyperparameter, so we report only settings that are explicitly recoverable from the project notes: the RADIO-style adaptor design, initialization from a \dinothree{} checkpoint when stated, the recorded teacher-weight codes, MIM ratio $0.3$ when enabled, and the ImageNet-22K training lengths.

\paragraph{Presentation style.}
To keep the paper focused, we separate stable observations from open issues. Tables~\ref{tab:main}-\ref{tab:challenge} summarize the representative numbers that support the main story; more speculative combinations are discussed only in the limitations section.

\begin{table}[t]
  \centering
  \caption{Main comparison across the project milestones. The best classification run and the best dense-recovery run are not the same model, which is exactly the tension that motivates \phis{}. Here \texttt{s}/\texttt{d} denote SigLIP/DINO loss weights.}
  \label{tab:main}
  \small
\setlength{\tabcolsep}{4pt}
\begin{tabular}{p{0.50\columnwidth}cccc}
\toprule
Model / setting & Patch & CLS & mIoU & mAcc \\
\midrule
\dinothree{} teacher & 79.7 & 85.0 & 48.5 & 61.0 \\
\method{} best cls (\texttt{s1d0.5}) & \textbf{83.2} & \textbf{85.2} & 46.5 & 58.1 \\
\method{} + \phis{} (IN1K) & 81.3 & 84.6 & \textbf{48.5} & \textbf{61.0} \\
\method{} + \phis{} (IN22K, \texttt{s1d1}) & 80.0 & 84.5 & 48.28 & 60.25 \\
\bottomrule
\end{tabular}

\end{table}

\begin{table}[t]
  \centering
  \caption{Adaptor and loss ablation. Splitting CLS and patch heads and assigning different losses to them produces the first major gain.}
  \label{tab:adaptor}
  \small
\setlength{\tabcolsep}{4pt}
\begin{tabular}{lcc}
\toprule
Adaptor / loss design & Patch & CLS \\
\midrule
Single adaptor, cosine only & 82.2 & 82.9 \\
Split adaptor, cosine + MSE & \textbf{82.7} & \textbf{84.1} \\
\dinothree{} teacher & 79.7 & 85.0 \\
\bottomrule
\end{tabular}

\end{table}

\begin{table}[t]
  \centering
  \caption{Initialization and teacher reweighting. Starting from \dinothree{} and rebalancing teacher losses moves the student past the teacher on ImageNet-1K kNN. Here \texttt{s}/\texttt{d} denote SigLIP/DINO loss weights.}
  \label{tab:init}
  \small
\setlength{\tabcolsep}{4pt}
\begin{tabular}{p{0.56\columnwidth}cc}
\toprule
Setting & Patch & CLS \\
\midrule
Split adaptor baseline & 82.7 & 84.1 \\
+ \dinothree{} init, \texttt{s1d1} & 82.4 & 84.8 \\
+ reweight, \texttt{s0.5d1} & 81.0 & 84.5 \\
+ reweight, \texttt{s1d0.5} & \textbf{83.2} & \textbf{85.2} \\
\bottomrule
\end{tabular}

\end{table}

\section{Results}
\label{sec:results}

\subsection{Classification improves through structure and teacher weighting}

The first phase of progress is mostly architectural. Table~\ref{tab:adaptor} shows that moving from a single adaptor with cosine-only supervision to split CLS/patch adaptors with cosine on CLS and MSE on patches lifts kNN from 82.2/82.9 to 82.7/84.1. Table~\ref{tab:init} then shows that initialization from \dinothree{} improves the CLS branch further, and explicit teacher reweighting with \texttt{s1d0.5} (SigLIP weight $1$, DINO weight $0.5$) yields the best classification setting: 83.2 patch kNN and 85.2 CLS kNN, slightly above the \dinothree{} teacher.

\begin{figure*}[t]
  \centering
  \includegraphics[width=0.95\textwidth]{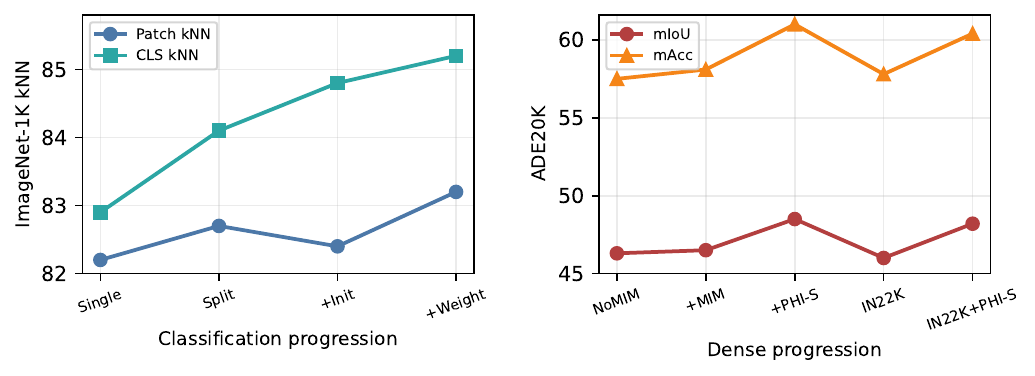}
  \caption{Project progression. Left: ImageNet-1K kNN improves through structural changes, initialization, and teacher reweighting. Right: ADE20K improves only modestly with MIM but is largely restored by \phis{}; the ImageNet-22K transition remains unstable.}
  \label{fig:progress}
\end{figure*}

This progression is visualized in Figure~\ref{fig:progress}. The important lesson is that the student does not become stronger by adding ingredients indiscriminately. Instead, the project improved by isolating the two token types and giving each branch a training signal that matches its role: directional alignment for the summary token and magnitude-sensitive regression for dense patches.

\begin{table}[t]
  \centering
  \caption{Dense recovery on ADE20K. MIM gives a small gain, but \phis{} is the key intervention that closes the dense-performance gap.}
  \label{tab:dense}
  \small
\setlength{\tabcolsep}{4pt}
\begin{tabular}{lcc}
\toprule
Setting & mIoU & mAcc \\
\midrule
No MIM, no \phis{} & 46.3 & 57.5 \\
+ MIM & 46.5 & 58.1 \\
+ MIM + \phis{} & \textbf{48.5} & \textbf{61.0} \\
\dinothree{} teacher & \textbf{48.5} & \textbf{61.0} \\
\bottomrule
\end{tabular}

\end{table}

\subsection{Dense performance is recovered by PHI-S}

Classification gains alone do not solve the project objective. Without additional balancing, the best ImageNet-1K recipe still trails \dinothree{} by a large margin on ADE20K. Table~\ref{tab:dense} shows the dense side of the story. Adding MIM improves mIoU/mAcc from 46.3/57.5 to 46.5/58.1, suggesting that masked prediction does help the patch representation. The much larger change comes from \phis{}, which restores segmentation all the way to 48.5/61.0, matching \dinothree{} on this benchmark.

This behavior is consistent with the geometry in Figure~\ref{fig:phis}. The student already has the capacity to absorb both teachers, but ordinary matching leaves dense supervision under-balanced. \phis{} acts less like a generic optimization trick and more like a targeted correction for teacher-statistics mismatch. In our current reading of the project, \phis{} is therefore the decisive ingredient for dense semantics, even though it does not define the best classification model.

There is also an important asymmetry hidden inside this result. In the tuned two-teacher recipe, MIM is mildly helpful. In isolation, however, it is far from universally benign. Table~\ref{tab:mim} shows that DINO-only distillation becomes unstable when MIM is added with the original low learning rate: patch/CLS kNN fall from 79.5/83.0 to 67.2/70.0. Raising the learning rate back to $5\times 10^{-4}$ or $7.5\times 10^{-4}$ recovers most of the lost performance. We therefore interpret MIM as a conditional regularizer whose effect depends strongly on the optimization regime and the teacher mixture, not as a universally positive add-on.

\begin{table}[t]
  \centering
  \caption{MIM sensitivity in DINO-only distillation. The auxiliary objective is not uniformly helpful; it becomes viable only after re-tuning optimization.}
  \label{tab:mim}
  \small
\setlength{\tabcolsep}{4pt}
\begin{tabular}{lcc}
\toprule
DINO-only setting & Patch & CLS \\
\midrule
MIM 0.0, LR $1.5\times 10^{-4}$ & 79.5 & 83.0 \\
MIM 0.3, LR $1.5\times 10^{-4}$ & 67.2 & 70.0 \\
MIM 0.3, LR $5\times 10^{-4}$ & 78.8 & 82.1 \\
MIM 0.3, LR $7.5\times 10^{-4}$ & 79.2 & 82.8 \\
\bottomrule
\end{tabular}

\end{table}

\subsection{Teacher balance exposes the global-dense trade-off}

The clearest internal trade-off in the progress log comes from the teacher-weight sweep. Increasing the \siglip{} weight consistently improves CLS-token recognition, but it also weakens segmentation because the student drifts away from the denser \dinothree{} representation. Table~\ref{tab:balance} makes this trend explicit: moving from \texttt{s1d1} (SigLIP weight $1$, DINO weight $1$) to \texttt{s1d0.1} (SigLIP weight $1$, DINO weight $0.1$) raises CLS kNN from 84.8 to 85.7 while mIoU drops from 47.6 to 42.0. The patch branch follows the same pattern, but less dramatically.

\begin{figure}[t]
  \centering
  \includegraphics[width=0.95\columnwidth]{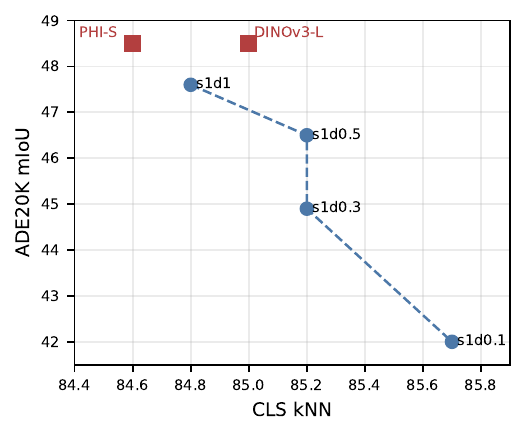}
  \caption{CLS kNN versus ADE20K mIoU for the teacher-weight sweep. Here \texttt{s}/\texttt{d} denote SigLIP/DINO loss weights. \phis{} improves the trade-off curve rather than simply selecting another point on the original sweep.}
  \label{fig:tradeoff}
\end{figure}

Figure~\ref{fig:tradeoff} provides another view of the same phenomenon. The raw weight sweep moves the model along a downward trade-off frontier between classification and dense prediction. The \phis{} point is interesting precisely because it does not continue that same line: it recovers dense performance to the \dinothree{} level while sacrificing much less classification than the teacher-only baseline. In other words, \phis{} does not remove the trade-off entirely, but it bends the frontier in a favorable direction.

\begin{table}[t]
  \centering
  \caption{Teacher-balance sweep on IN-1K and ADE20K. Heavier \siglip{} weighting benefits classification but erodes dense performance. Here \texttt{s}/\texttt{d} denote SigLIP/DINO loss weights.}
  \label{tab:balance}
  \small
\setlength{\tabcolsep}{3pt}
\begin{tabular}{lcccc}
\toprule
Weight (\texttt{s}=\siglip{}, \texttt{d}=\dinothree{}) & Patch & CLS & mIoU & mAcc \\
\midrule
\texttt{s1d1} & 82.4 & 84.8 & 47.6 & 59.2 \\
\texttt{s1d0.5} & 83.2 & 85.2 & 46.5 & 58.1 \\
\texttt{s1d0.3} & 83.2 & 85.2 & 44.9 & 56.6 \\
\texttt{s1d0.1} & 83.1 & \textbf{85.7} & 42.0 & 53.7 \\
\texttt{s1d0.5} + \phis{} & 81.3 & 84.6 & \textbf{48.5} & \textbf{61.0} \\
\bottomrule
\end{tabular}

\end{table}

\subsection{Scaling and teacher expansion remain unresolved}

Table~\ref{tab:challenge} summarizes the less positive part of the story. Moving from ImageNet-1K to ImageNet-22K does not produce a consistent gain. Even after fixing the \phis{} statistics computation for ImageNet-22K and reducing the training length to 30 epochs, the ImageNet-22K student remains below the best ImageNet-1K+\phis{} result. Reweighting the teachers from \texttt{s1d0.5} (SigLIP weight $1$, DINO weight $0.5$) to \texttt{s1d1} (SigLIP weight $1$, DINO weight $1$) helps segmentation, but the overall picture still suggests that more data alone is not enough.

The same pattern appears when adding more teachers. SAM3 slightly improves neither classification nor segmentation in the current setup, and adding HOG-like features degrades both. These negative results matter because they constrain the story: the current project does not yet support the claim that simply adding more specialist teachers will continue to improve the student. What the evidence supports instead is a narrower but clearer claim: within a carefully tuned two-teacher system, \method{} can surpass the teacher on classification and recover the teacher on dense prediction, but larger-scale aggregation still requires better interference control.

One concrete diagnosis from the progress log is that scaling issues are partly statistical rather than purely architectural. Table~\ref{tab:in22kstats} compares three ways of obtaining \phis{} statistics for ImageNet-22K. Reusing ImageNet-1K statistics gives a reasonable baseline; recomputing them with repeated samples from the ImageNet-22K loader severely damages the CLS branch; recomputing them correctly with a single full pass restores the classification metric. This bug fix improves our confidence that the remaining ImageNet-22K gap is a real modeling problem rather than a bookkeeping artifact.

The same section of the log also reports that, once \phis{} is active, the ImageNet-22K mIoU curve saturates early: 30, 35, 40, and 50 epochs all stay within a narrow 47.8--48.2 range. This suggests that the current two-teacher recipe is not simply undertrained on ImageNet-22K. Instead, it appears to converge to a different trade-off point than the ImageNet-1K recipe, supporting our view that better scaling will require changes in objective design or teacher coordination rather than simply training longer on the same recipe.

\begin{table}[t]
  \centering
  \caption{Scaling and challenge summary. Larger-scale distillation and extra teachers still show interference, even when they improve one side of the trade-off. Here \texttt{s}/\texttt{d} denote SigLIP/DINO loss weights.}
  \label{tab:challenge}
  \small
\setlength{\tabcolsep}{3pt}
\begin{tabular}{p{0.50\columnwidth}cccc}
\toprule
Setting & Patch & CLS & mIoU & mAcc \\
\midrule
IN22K, no \phis{} (\texttt{s1d0.5}) & 81.9 & 84.5 & 46.0 & 57.8 \\
IN22K + \phis{} (\texttt{s1d0.5}) & 80.6 & 84.3 & 47.40 & 58.94 \\
IN22K + \phis{} (\texttt{s1d1}) & 80.0 & 84.5 & \textbf{48.28} & \textbf{60.25} \\
+ SAM3 & 84.0 & 79.4 & 47.59 & 59.82 \\
+ SAM3 + HOG & 78.9 & 69.9 & 47.31 & 59.26 \\
\bottomrule
\end{tabular}

\end{table}

\begin{table}[t]
  \centering
  \caption{PHI-S statistics diagnostics on ImageNet-22K. The corrected one-pass statistics recover the classification branch that was damaged by repeated-sample estimation.}
  \label{tab:in22kstats}
  \small
\setlength{\tabcolsep}{4pt}
\begin{tabular}{lcc}
\toprule
PHI-S statistics source & Patch & CLS \\
\midrule
Reused IN1K statistics & 80.8 & 84.3 \\
IN22K repeated-sample estimate & 80.1 & 78.4 \\
IN22K one-pass fixed estimate & 80.6 & 84.3 \\
\bottomrule
\end{tabular}

\end{table}

\begin{table}[t]
  \centering
  \caption{ImageNet-22K epoch sweep with \phis{}. Here \texttt{s}/\texttt{d} denote SigLIP/DINO loss weights. Once the corrected statistics are used, the dense metric saturates early rather than improving monotonically with longer training.}
  \label{tab:in22kepochs}
  \small
\setlength{\tabcolsep}{4pt}
\begin{tabular}{lcc}
\toprule
Epochs (\texttt{s1d0.5}, \texttt{s}=\siglip{}, \texttt{d}=\dinothree{}) & mIoU & mAcc \\
\midrule
30 & 47.84 & 59.84 \\
35 & 47.80 & 59.71 \\
40 & \textbf{48.24} & \textbf{60.01} \\
50 & 47.95 & 59.78 \\
\bottomrule
\end{tabular}

\end{table}

\section{Discussion}

The current project can be read as a study of representation interference. \siglip{} and \dinothree{} are not merely different teachers with different strengths; they also shape different geometry in the student space. The split-head design matters because it prevents the CLS pathway from dictating the behavior of every patch token. The weight sweep matters because it reveals that there is no single teacher ratio that is simultaneously optimal for recognition and segmentation. And \phis{} matters because it acts on feature statistics rather than only on scalar teacher weights.

This view also clarifies why the negative results are informative. The failure of SAM3 and HOG to improve the current recipe does not imply that dense auxiliary teachers are useless. Instead, it suggests that the present pipeline has limited capacity to coordinate more than two strong objectives without additional routing, teacher selection, or token-level specialization.

Several practical limitations remain. Our evaluation is restricted to linear-probe segmentation and kNN classification; downstream tasks such as detection or depth estimation may reveal different trade-off landscapes. All experiments use a ViT-B student; larger capacities could absorb more teachers before interference becomes dominant. Finally, the \phis{} statistics are computed offline; an online variant might adapt more gracefully as the student evolves.

\section{Conclusion}

We have presented \method{}, a multi-teacher distillation recipe that improves dense semantic quality in AM-RADIO-style training by combining three ingredients: split CLS/patch heads, \phis{}-based feature normalization, and masked image modeling as an auxiliary regularizer. Experiments on ImageNet-22k show that these additions recover dense segmentation performance lost under naive \siglip{}+\dinothree{} distillation while maintaining strong classification and VLM transfer. Our ablations reveal that no single scalar teacher ratio is simultaneously optimal for recognition and segmentation, making feature-space interventions like \phis{} essential complements to loss balancing. We hope this analysis provides a useful reference for practitioners building general-purpose vision encoders from heterogeneous teachers.

\FloatBarrier
\bibliographystyle{plainnat}
\bibliography{references}

\end{document}